\documentclass[acmtog]{acmart}
\AtBeginDocument{%
  }

\acmSubmissionID{2256}

\usepackage{multirow}
\usepackage{booktabs}
\usepackage{makecell}
\usepackage{diagbox}
\usepackage{cleveref}
\usepackage{enumerate}
\usepackage{xcolor}
\usepackage{pifont}
\usepackage{enumitem}

\copyrightyear{2026}
\acmYear{2026}
\setcopyright{cc}
\setcctype{by-nc-nd}
\acmConference[SA Conference Papers '26]{SIGGRAPH Asia 2026 Conference Papers}{December 01--04, 2026}{Kuala Lumpur, Malaysia}
\acmBooktitle{SIGGRAPH Asia 2026 Conference Papers (SA Conference Papers '26), December 01--04, 2026, Kuala Lumpur, Malaysia}
\acmDOI{10.1145/3829340.3842336}
\acmISBN{979-8-4007-2842-6/2026/12}

\definecolor{cmarkgreen}{RGB}{0, 180, 0}
\definecolor{xmarkred}{RGB}{200, 0, 0}

\definecolor{revisionblue}{RGB}{0, 92, 175}

\definecolor{cameraready}{RGB}{0, 92, 175}

\newcommand{\cmark}{\textcolor{cmarkgreen}{\ding{51}}} 
\newcommand{\xmark}{\textcolor{xmarkred}{\ding{55}}}   

\begin{document}

\title{InterMASH: A Unified Geometric Representation for Grasp Synthesis}

\begin{teaserfigure}
  \centering
  \includegraphics[width=1.0\linewidth]{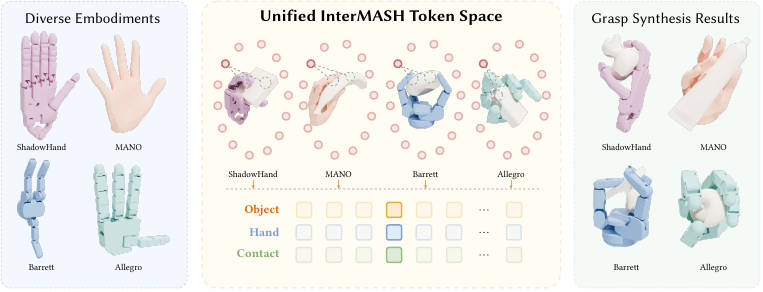}
  \caption{Overview of InterMASH. Diverse human and robotic hand embodiments are represented in a shared anchor-indexed InterMASH token space, where corresponding anchors encode local object geometry, hand geometry, and contact information. Operating in this unified interaction space enables cross-embodiment grasp synthesis for ShadowHand, MANO, Barrett, and Allegro.}
  \label{fig:teaser}
\end{teaserfigure}

\author{Xuanze Yang}
\email{young_xz@mail.ustc.edu.cn}
\orcid{0009-0002-8298-1668}
\affiliation{%
  \institution{University of Science and Technology of China}
  \city{Hefei}
  \country{China}
}

\author{Yumeng Liu}
\email{lym29@mail.ustc.edu.cn}
\orcid{0000-0003-1866-6727}
\authornote{Corresponding Author: Yumeng Liu (lym29@mail.ustc.edu.cn)}
\affiliation{%
  \institution{University of Science and Technology of China}
  \city{Hefei}
  \country{China}
}

\author{Haiyang Xin}
\email{xhy2878@mail.ustc.edu.cn}
\orcid{0009-0009-7486-3267}
\affiliation{%
  \institution{University of Science and Technology of China}
  \city{Hefei}
  \country{China}
}

\author{Changhao Li}
\email{lch0510@mail.ustc.edu.cn}
\orcid{0000-0003-0850-8987}
\affiliation{%
  \institution{University of Science and Technology of China}
  \city{Hefei}
  \country{China}
}

\author{Haowei Shen}
\email{sh030705@mail.ustc.edu.cn}
\orcid{0009-0001-4324-799X}
\affiliation{%
  \institution{University of Science and Technology of China}
  \city{Hefei}
  \country{China}
}

\author{Kai Xu}
\email{kevin.kai.xu@gmail.com}
\orcid{0000-0002-9054-0216}
\affiliation{%
\institution{Jiangsu Key Laboratory of AI for Industries and Institute of AI for Industries Chinese Academy of Sciences}
  \city{Nanjing}
  \country{China}
}

\author{Ligang Liu}
\email{lgliu@ustc.edu.cn}
\orcid{0000-0003-4352-1431}
\affiliation{%
  \institution{University of Science and Technology of China}
  \city{Hefei}
  \country{China}
}

\author{Ruizhen Hu}
\email{ruizhen.hu@gmail.com}
\orcid{0000-0002-6798-0336}
\affiliation{%
  \institution{Shenzhen University}
  \city{Shenzhen}
  \country{China}
}


\begin{abstract}
Grasp synthesis aims to generate stable and physically plausible hand--object interactions, and has become a fundamental problem in both human hand modeling and robotic manipulation. However, a unified representation across human and robotic hands is still lacking, mainly due to differences in hand morphology and surface modeling. Prior methods typically rely on either contact maps or dense implicit descriptors to represent interaction, but these representations are often incomplete or computationally expensive and redundant. We propose InterMASH, a unified geometric representation that establishes cross-embodiment correspondence using sphere-fixed anchors. At each anchor, low-degree spherical harmonics compactly encode local hand geometry, object geometry, and contact, forming an explicit and interpretable token sequence. Building on this natively tokenized structure, we introduce a conditional Diffusion Transformer that operates directly in the proposed InterMASH representation space and jointly generates hand geometry and contact, improving consistency and physical plausibility. Our method achieves competitive performance with state-of-the-art methods on key physical feasibility metrics in a large-scale ShadowHand benchmark, supports joint training across multiple hands, and shows that cross-embodiment fine-tuning with human grasp data can improve robotic grasp success and diversity. Project page is available at \url{https://inter-mash.github.io/}.
\end{abstract}




\maketitle

\section{Introduction}
Generating natural and functional grasps is a central challenge in computer graphics and embodied intelligence, playing a crucial role in both digital avatar manipulation~\cite{zhang2021manipnet,zhang2024hand,zhang2025bimart} and robotic tasks~\cite{zhang2024graspxl}. Humans exhibit highly effective and adaptable manipulation skills, yet transferring such versatility to robotic systems remains difficult due to the substantial morphological and geometric differences between human and robotic hands. At the same time, grasp synthesis is inherently a high-dimensional, constrained generation problem that must satisfy both geometric and physical requirements across diverse objects~\cite{liu2021synthesizing,li2022gendexgrasp}. A key challenge is therefore to develop a grasp representation that is both expressive enough to capture the local geometry of hand-object interaction and abstract enough to generalize across different embodiments.


Existing works have explored interaction representations ranging from object-centric contact maps~\cite{jiang2021graspTTA,yang2021cpf,li2022gendexgrasp,wei2025afforddexgrasp} to dense geometric representations such as point-to-point distance representations~\cite{wei2024d} and SDF-based fields~\cite{ye2024g}. Recent studies on deep 3D shape generation have highlighted that representation design plays a critical role in balancing geometric fidelity, compactness, and generation efficiency~\cite{xu2023survey,wang2025diffusion3d}. However, contact maps mainly encode whether and where contact occurs, without explicitly representing the hand geometry involved in the grasp, while dense geometric representations describe hand-object interactions through pairwise relations or spatial fields that include redundant information irrelevant to the grasp. Consequently, existing interaction representations either lack sufficient geometric description for fine-grained grasp modeling or fail to provide a compact basis for scalable generative modeling.


Inspired by BPS~\cite{prokudin2019efficient} and MASH~\cite{li2025mash}, we propose \textbf{InterMASH}, a unified interaction representation that jointly achieves compactness and expressiveness while enabling grasp synthesis across different hand morphologies. From BPS, we adopt the idea of placing a fixed set of points in space as a shared indexing scheme. In our case, these points are instantiated as anchors on a sphere centered at the object, establishing a morphology-agnostic reference frame under which interactions from different hands map to a common token space. From MASH, we adopt low-degree spherical harmonic (SH) coefficients as a compact local descriptor at each anchor, replacing BPS's scalar distance with a richer encoding that captures the local hand surface, object surface, and contact map. This combination gives the representation three key characteristics: a \textbf{unified} token space shared across different hand morphologies, a \textbf{compact} encoding with only a handful of SH coefficients per anchor, and an \textbf{expressive} description of fine-grained local geometry rather than a single scalar.

However, directly adapting these two representations to hand-object interaction is non-trivial. First, with fixed anchors, there is no built-in semantic correspondence between anchor indices and hand surface regions---the same anchor may map to unrelated spatial regions across different hands. We resolve this by first establishing a MANO reference patch ordering and then aligning robotic-hand templates to MANO with keypoint-distance signatures, yielding a consistent semantic ordering that transfers to all grasp instances. Second, jointly fitting hand and object SH coefficients under shared anchors is unstable under direct optimization; we adopt a coarse-to-fine schedule that progressively increases the SH degree, stabilizing convergence.

Building on InterMASH, we further propose a generative framework based on a conditional Diffusion Transformer (cDiT)~\cite{peebles2023scalable}, which operates directly in this anchor-indexed interaction space. Unlike prior approaches that predict object contact as a separate intermediate signal for downstream optimization~\cite{jiang2021graspTTA,xu2023unidexgrasp}, our framework jointly synthesizes hand geometry and contact information in the same compact representation. This joint modeling promotes mutual consistency between shape and contact, improves physical plausibility by reducing penetration, and enables grasp generation across diverse embodiments within a unified generative framework.

Our design choices are supported by extensive empirical validation. Ablation studies confirm that the SH-based local descriptor achieves a better compactness--expressiveness trade-off than alternatives such as scalar distances and learned patch embeddings. On a large-scale dexterous grasping benchmark, our full framework achieves state-of-the-art performance on key physical feasibility metrics. We further show that InterMASH supports cross embodiment synthesis, achieving the best success rate on Barrett while remaining competitive on ShadowHand. Finally, our experiments demonstrate that the InterMASH representation enables joint training on human and robotic hand data, which in turn benefits robotic grasp generation.

In summary, our work makes the following contributions:
\begin{itemize}[topsep=4pt, itemsep=1pt, parsep=0pt]
\item We introduce InterMASH, a unified hand-object interaction representation that combines fixed anchors with compact spherical harmonic (SH) local descriptors, establishing consistent cross-embodiment correspondence while preserving fine-grained local interaction geometry.

\item We develop a conditional DiT-based generative framework that operates directly in the InterMASH token space and jointly synthesizes hand geometry and contact, improving physical plausibility and shape-contact consistency.

\item We demonstrate that InterMASH provides a unified representation that enables joint training on data from different hand morphologies, and that integrating human grasp data can improve robotic grasp success rate and diversity.
\end{itemize}

\section{Related Work}
\paragraph{Grasp Synthesis}
Grasp synthesis has been studied for both human and robotic hands. Although the two settings share the goal of generating plausible, functional, and diverse grasps, they have largely evolved with distinct data sources, representations, and embodiment-specific models.
For human grasp synthesis, early works relied on optimization-based methods to satisfy kinematic and geometric constraints~\cite{liu2021synthesizing}. Related transfer-based approaches reuse existing hand-object interactions by retargeting them to new object categories~\cite{Zhou2022TOCHSO,wang2026parthoi}. However, such methods require source interactions and corresponding semantic object parts, while their optimization-based retargeting limits scalability and grasp diversity. 
Recent data-driven approaches for human grasp synthesis have achieved significant progress~\cite{karunratanakul2020grasping,wu2025fastgrasp,zhang2025bimart,cao2026jointdiffusionuniversalhandobject} that can yield natural, human-like grasps thanks to high-quality motion-capture datasets~\cite{hampali2020honnotate,hampali2022keypointtransformer,GRAB-2020,YangCVPR2022OakInk,fan2023arctic}.
However, human hand-object interactions, whether captured from real data or synthesized by generative models, cannot be directly used for robotic grasp learning due to differences in hand morphology and kinematics, requiring an additional retargeting step. This additional processing introduces substantial computational overhead when generating interactions at scale.
In robotics, large-scale dexterous grasp datasets are commonly constructed through simulation-based pipelines~\cite{li2022gendexgrasp,wang2022dexgraspnet}. Although scalable, such datasets primarily emphasize physical feasibility and grasp stability rather than human-like dexterity. Consequently, models trained on these data, such as UniDexGrasp~\cite{xu2023unidexgrasp,Wan_2023_ICCV}, may produce stable but less natural grasps. An alternative is to collect human-aligned robotic grasps through human-to-robot retargeting~\cite{qin2023anyteleop} or tele-operation~\cite{liu2024realdex}. Recent approaches such as DexGrasp Anything~\cite{zhong2025dexgrasp} demonstrate that training on such human-like robotic grasp data can substantially improve grasp generation quality. However, this paradigm exploits human grasping priors only indirectly, after costly retargeting or tele-operation for a specific robotic hand, making data collection embodiment-specific and difficult to scale. This gap motivates a unified framework that transfers human-like interaction priors across diverse hand embodiments. 



\begin{figure*}[h]
    \centering
    \includegraphics[width=1.0\linewidth]{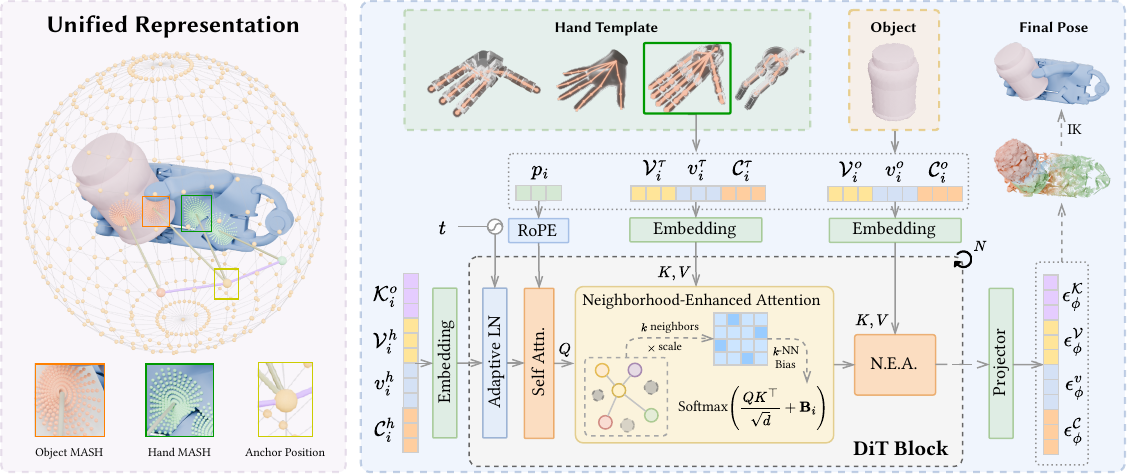}
    \caption{Overview of InterMASH: We first define InterMASH to uniformly represent the geometric relationship between hand and object. Next, we introduce a conditional diffusion model to synthesize hand geometry conditioned on the hand template and object.}
    \label{fig:overview}
\end{figure*}

\paragraph{Geometric Representations for Interaction}
Geometric representations are fundamental for modeling hand-object interactions. Early methods use object-centric contact maps~\cite{jiang2021graspTTA,li2022gendexgrasp,yang2021cpf}, which predict contact regions on the object but ignore hand geometry, leading to kinematic ambiguities. Recent research has shifted toward interaction-centric representations that explicitly capture the spatial relationship between hand and object. \citet{she2024learning} propose to use Interaction Bisector Surfaces (IBS)~\cite{zhao2014indexing} to provide a detailed description of the hand-object interface, but require computationally intensive spatial partitioning. \citet{wei2024d} employ an implicit point-to-point distance representation to enable cross-embodiment grasp synthesis; however, this approach leads to high-dimensional descriptors, and the hand point clouds extracted from the implicit distance field are often noisy and lack geometric fidelity.  Similarly, \citet{ye2024g} introduce skeletal distance fields to jointly model hand and object geometry. However, this volumetric representation leads to high memory use and a loss of fine detail. Concurrently, $\mathcal{T(R,O)}$ Grasp~\cite{fei2025t} increases efficiency using graph diffusion on robotic links, but is limited to rigid-body systems and cannot model continuous hand deformations, making it incompatible with human interaction priors. These limitations motivate the need for expressive, compact, and explicit geometric representations for cross-embodiment grasp synthesis.

\section{Preliminary}

In this section, we briefly review the components of MASH (Masked Anchored Spherical Distances)~\cite{li2025mash} that are most relevant to our hand-object interaction formulation. MASH is particularly suitable for our setting because it represents a surface as a collection of aligned local patches, which naturally supports compact geometric encoding and token-wise processing.

Given a 3D shape $\mathcal{S} \subset \mathbb{R}^3$, MASH represents its surface using a set of \emph{anchors} $\{\mathcal{A}_i\}_{i=1}^n$. Each anchor acts as a virtual viewpoint centered at $p_i$ and captures a localized surface patch around that observation center. For an anchor $\mathcal{A}_i$, rays are cast from $p_i$ along spherical directions $(\theta, \phi)$, and the distance to the first visible surface intersection defines a spherical distance function $d_i(\theta, \phi)$.

To obtain a compact and differentiable parameterization, each spherical distance function is approximated by spherical harmonics:
\begin{equation*}
    d_i(\theta, \phi) = \sum_{l=0}^{L} \sum_{m=-l}^{l} C^{(i)}_{l,m} \, Y_l^m(\theta, \phi),
\end{equation*}
where $Y_l^m$ denotes the SH basis functions and $C^{(i)}_{l,m}$ are the corresponding coefficients.

Directly approximating a full spherical distance function is challenging because visibility changes introduce occlusions and discontinuities. MASH addresses this issue by introducing a \emph{vision mask} that restricts each anchor to a localized angular support. Concretely, the mask is defined by a generalized view cone with anisotropic boundary $\alpha_i(\phi)$, which limits the valid range of $\theta$ for each azimuth $\phi$. This visibility-aware masking enforces locality and makes low-order SH sufficient to approximate each local patch in a compact and stable manner.

Each anchor is therefore parameterized as $\mathcal{A}_i = \{p_i, v_i, \mathcal{C}_i, \mathcal{V}_i\}$, where $p_i$ is the anchor position, $v_i$ defines its local orientation, $\mathcal{C}_i$ denotes the SH coefficients, and $\mathcal{V}_i$ represents the vision-mask parameters. The full shape is reconstructed as the union of these anchored local patches.

This explicit, local, and patch-based structure makes MASH a strong starting point for interaction modeling, as it decomposes geometry into a set of consistent local primitives rather than a dense global descriptor. In the following section, we extend this single-shape representation to a joint hand-object formulation with shared anchors, cross-instance correspondence, and contact-aware geometric encoding for dexterous grasp synthesis.

\section{Methodology}
\label{sec:method}
We formulate grasp synthesis as geometric conditional generation: given a target object $O$ and a template hand geometry $H_{\tau}$ of hand type $\tau$, we model the conditional distribution $p(H\mid O, H_{\tau})$ of valid grasping hand geometries. To address this problem, we propose a three-stage pipeline. First, we represent hand-object interaction using InterMASH (\autoref{sec:mash_hoi_rep}), a shared anchor-based representation that encodes local hand geometry, object geometry, and contact in a unified token space. A conditional Diffusion Transformer (DiT) then generates grasp-specific hand geometry and contact directly in this space (\autoref{sec:diffusion}). Finally, we decode the generated geometry into an articulated hand pose through inverse kinematics (IK, \autoref{sec:ik}). An overview of the pipeline is shown in \autoref{fig:overview}.

\begin{figure}[t]
    \centering
    \includegraphics[width=1.0\linewidth]{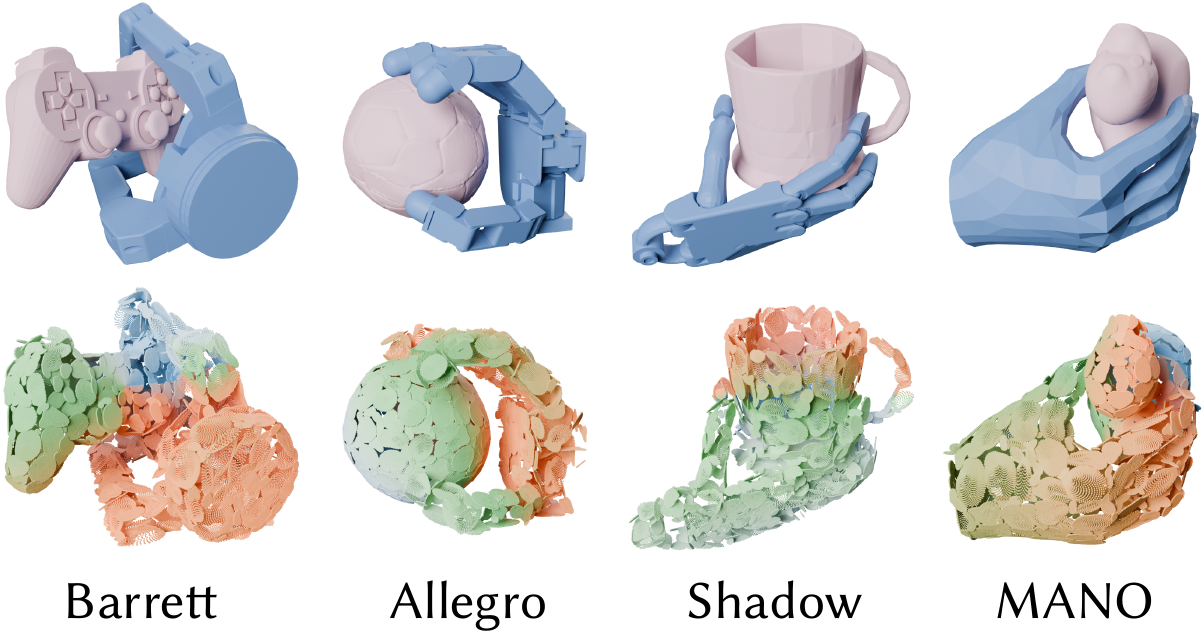}
    \caption{Visualization of the original meshes (top) and their corresponding InterMASH representations (bottom) across different hand models, including Barrett, Allegro, Shadow and MANO.}
    \label{fig:mesh_and_mash}
\end{figure}

\subsection{InterMASH Representation}
\label{sec:mash_hoi_rep}

Inspired by recent progress in structured 3D representation learning, where feature organization is critical for preserving geometric details~\cite{su2023point}, we design anchor-based tokens to explicitly encode local hand-object structures by combining two prior representations: BPS~\cite{prokudin2019efficient} and MASH~\cite{li2025mash}. From BPS, we adopt the idea of a fixed set of points serving as a shared spatial indexing scheme, which establishes consistent correspondence across instances and hand morphologies; however, unlike BPS, which stores a single nearest-surface displacement per basis point, each InterMASH anchor encodes a full local surface patch via low-degree SH coefficients. From MASH, we retain the anchor-based parameterization (SH coefficients and vision mask) for compact local geometry, but with two key changes: anchors are fixed in a shared object-centered reference frame rather than optimized per instance, and each anchor jointly encodes hand geometry, object geometry, and contact instead of a single shape.

While effective for single-shape reconstruction, MASH's instance-specific anchors do not guarantee semantic consistency across different hand-object instances. For grasp synthesis, however, the model must compare and generate interactions across diverse objects and hand morphologies in a shared representation space.

To address this issue, we extend MASH to a joint hand-object representation with shared fixed anchors and explicit cross-instance correspondence. This design turns each anchor into a semantically meaningful local token, making the representation suitable for both cross-embodiment training and transformer-based generation. \autoref{fig:mesh_and_mash} shows examples of meshes and their corresponding InterMASH reconstructions across different hand models.

\paragraph{Joint Anchor Representation of Hand and Object}
As shown in \autoref{fig:overview}, we place a shared set of anchors $\{\mathcal{A}_i\}_{i=1}^n$ for each hand-object pair. Each anchor jointly encodes a local hand patch and a local object patch: $\mathcal{A}_i = \left\{ p_i,\, v_i^h,\, \mathcal{C}_i^h,\, \mathcal{V}_i^h,\, v_i^o,\, \mathcal{C}_i^o,\, \mathcal{V}_i^o \right\}$, where \(p_i\) is the anchor position, \(v\) defines local orientation, \(\mathcal{C}\) denotes the SH coefficients, and \(\mathcal{V}\) represents the vision-mask parameters; the superscripts \(h\) and \(o\) indicate the hand and object, respectively. Compared with separate shape representations, this joint parameterization directly captures the local geometry of both interacting surfaces within a single token.

To make these tokens consistent across instances, we fix the anchor positions on a sphere centered at the object and place them using Fibonacci sampling. We empirically set the sphere radius to $r=0.2$, which encloses all objects and hands in our data. Beyond providing uniform spatial coverage, these fixed anchors define an embodiment-agnostic reference frame, so that the same anchor index refers to comparable spatial regions across different hand morphologies. Details of the Fibonacci construction are provided in the supplementary material.

\begin{figure}
    \centering
    \includegraphics[width=1.0\linewidth]{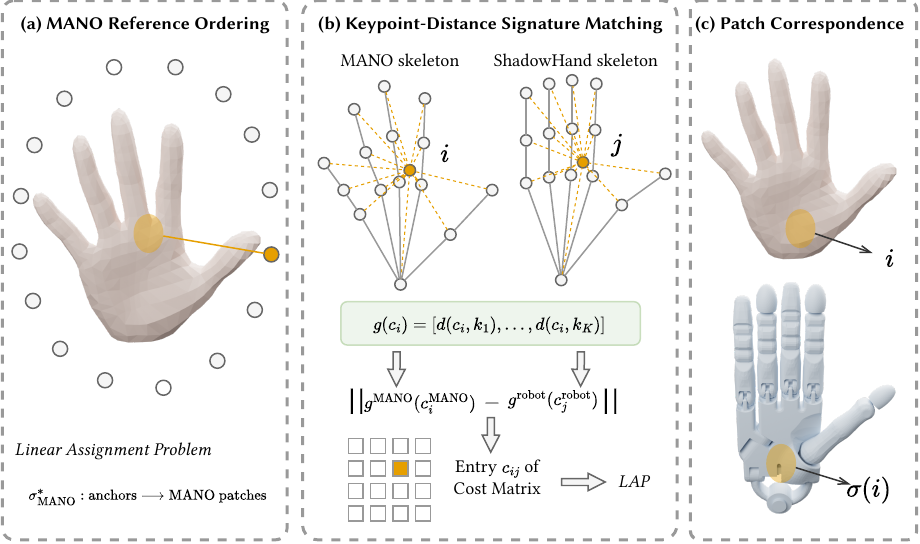}
    \caption{Cross-hand correspondence. Robotic-hand patches are aligned to a MANO reference via LAP-based matching in keypoint-distance space, establishing consistent anchor semantics across hand types.}
    \label{fig:anchor-patch-map}
\end{figure}

\paragraph{Consistent Anchor-Patch Correspondence} Fixed anchors alone do not define a semantic ordering over hand surface regions. A direct spatial assignment can associate the same anchor index with unrelated parts across different hand morphologies, which would make mixed-hand training ambiguous. We therefore build all hand-template patch orderings through a MANO-centered correspondence, as illustrated in \autoref{fig:anchor-patch-map}. First, we assign MANO template patches to the fixed anchors by solving an LAP between anchor positions and MANO patch centers, obtaining the reference ordering. For each robotic hand template $\tau$, we then align its patches to the MANO patches in a keypoint-distance feature space.

Specifically, we define homologous keypoints on each template hand and represent a patch center $c_i^{\tau}$ by a skeleton-based distance signature
\begin{equation*}
\mathbf{g}^{\tau}(c_i^{\tau}) =
\left[d_{\tau}(c_i^{\tau}, k_1^{\tau}), \ldots, d_{\tau}(c_i^{\tau}, k_K^{\tau})\right],
\end{equation*}
where $\{k_j^{\tau}\}_{j=1}^{K}$ are the keypoints of hand type $\tau$ and $d_{\tau}$ is a geodesic-like distance along the hand skeleton. We find the permutation from the robotic template to the MANO reference by solving
\begin{equation*}
\pi_{\tau}^* = \operatorname*{argmin}_{\pi \in S_n}
\sum_{i=1}^{n}
\left\|
\bar{\mathbf{g}}^{\mathrm{MANO}}(c_i^{\mathrm{MANO}})
-
\bar{\mathbf{g}}^{\tau}(c_{\pi(i)}^{\tau})
\right\|_2^2,
\end{equation*}
where $\bar{\mathbf{g}}$ denotes the normalized signature. This one-time template alignment maps ShadowHand, Barrett, and Allegro patches into the MANO patch order, so the same anchor index refers to a semantically corresponding hand region across embodiments. The resulting correspondence is transferred to all grasp instances through barycentric interpolation and is used by both InterMASH fitting and the IK stage in \autoref{sec:ik}. Implementation details are provided in the supplementary material.

\paragraph{Staged Optimization of SH Parameters}
Accurate reconstruction requires fitting the SH coefficients of each local patch. In practice, directly optimizing all coefficients, especially the high-frequency terms, leads to unstable early-stage convergence. We therefore adopt a coarse-to-fine schedule: optimization begins with degree $d_{\mathrm{sh}}=0$ and progressively increases the SH degree to $d_{\max}=2$ once the reconstruction loss stabilizes. This staged fitting process improves optimization stability and yields more accurate hand-object reconstructions, as validated in the supplementary material.

\paragraph{Compact Representation for Object Contact Map}
In addition to local hand and object geometry, we attach a compact contact descriptor to each anchor. Rather than representing contact densely on the object surface, we approximate the local contact field using a truncated SH expansion with only four coefficients:
\begin{equation*}
    \mathcal{K}_i=\sum_{l=0}^1\sum_{m=-l}^l C_{l, m}^{\left(i\right)}Y_l^m\left(\theta, \phi\right),
\end{equation*}
which encodes the spatial variation of contact strength around the corresponding object patch. Following~\cite{jiang2021graspTTA}, the contact strength at an object point \(P_{\mathrm{obj}}\) is defined as
\[
\mathcal{K}(P_{\mathrm{obj}}) = 1 - 2\left( \mathrm{Sigmoid}\left(100 \cdot D(P_{\mathrm{obj}})\right) - 0.5 \right),
\]
where \(D(P_{\mathrm{obj}})\) denotes the nearest distance between \(P_{\mathrm{obj}}\) and the hand point cloud. 
Together, the hand geometry, object geometry, and contact descriptors define the InterMASH tokens used by our generative model. The contact descriptor serves as auxiliary supervision during training and is not required at inference.

\subsection{Scalable Generative Model Design}
\label{sec:diffusion}
Each InterMASH anchor corresponds to one low-dimensional token, making the representation directly compatible with transformer-based generation. We therefore cast grasp synthesis as conditional sequence modeling in the InterMASH space and adopt a Diffusion Transformer (DiT)~\cite{peebles2023scalable} as the generative backbone. Operating directly on these compact tokens is substantially more efficient than modeling dense implicit interaction fields.

\paragraph{Model Framework}
The DiT predicts grasp-specific hand geometry and contact conditioned on the object geometry and template-hand geometry. Following the DDPM framework~\cite{ddpm}, we define the clean state $f_0$ as the anchor-wise feature sequence to be generated:
\begin{equation*}
f_0 = \left\{ v_i^{h}, \mathcal{C}_i^{h}, \mathcal{V}_i^{h}, \mathcal{K}_i \right\}_{i=1}^n.
\end{equation*}
The generated variables therefore consist of the hand-side MASH parameters and the object-side contact descriptor at each anchor. During the forward process, Gaussian noise is progressively added to $f_0$ over $T$ timesteps:
\begin{equation*}
q(f_t \mid f_0) = \mathcal{N}\left(f_t; \sqrt{\bar{\alpha}_t} f_0, (1-\bar{\alpha}_t)\mathbf{I}\right),
\end{equation*}
where $\bar{\alpha}_t = \prod_{s=1}^{t} (1-\beta_s)$. The model is trained to predict the added noise $\epsilon$ from noisy feature $f_t$, timestep $t$, and conditioning feature $\mathbf{c}$:
\begin{equation*}
\mathcal{L}_{\mathrm{recon}} = \mathbb{E}_{f_0,t,\epsilon}\left[\left\|\epsilon - \epsilon_{\phi}(f_t, \mathbf{c}, t)\right\|_2^2\right].
\end{equation*}

The reverse process recovers $f_0$ from a noisy state $f_t$ at timestep $t$, conditioned on the object geometry and template-hand geometry:
\begin{equation*}
\mathbf{c} = \left\{
v_i^o, \mathcal{C}_i^o, \mathcal{V}_i^o, v_i^{\tau}, \mathcal{C}_i^{\tau}, \mathcal{V}_i^{\tau}
\right\}_{i=1}^n.
\end{equation*}

\begin{figure}[t]
    \centering
    \includegraphics[width=0.85\linewidth]{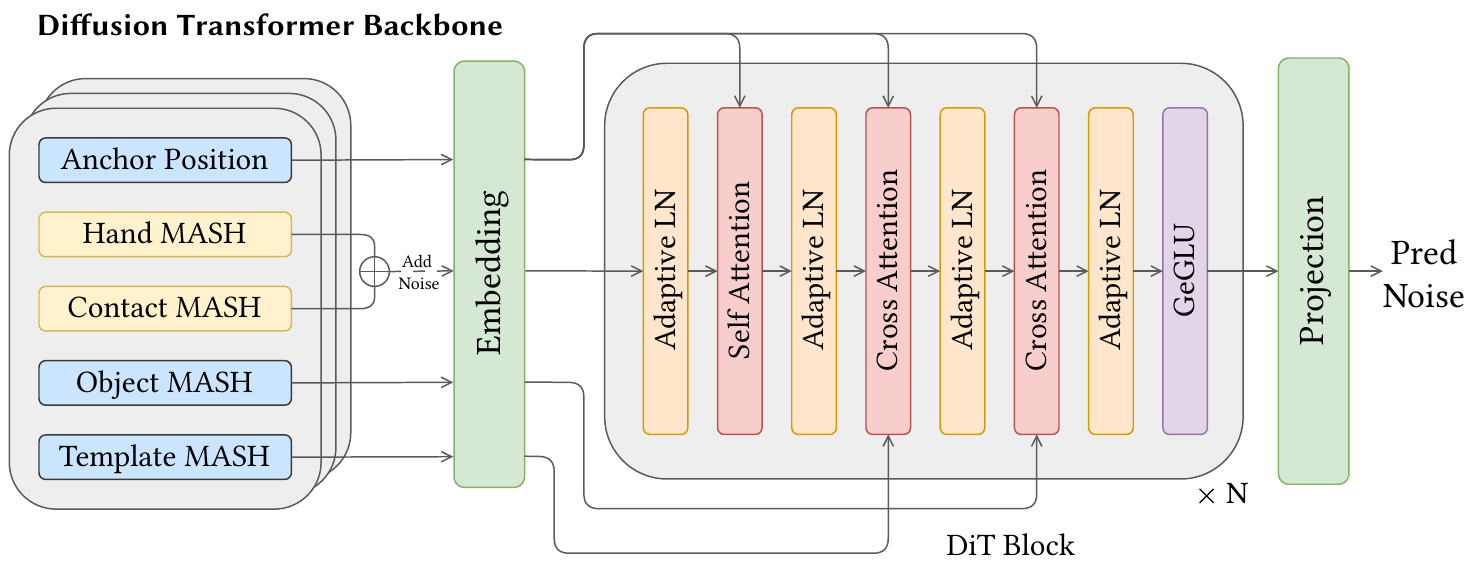}
    \caption{Diffusion Transformer backbone for InterMASH generation.}
    \label{fig:backbone}
\end{figure}

The model consists of $N$ stacked DiT blocks, as shown in \autoref{fig:backbone}. Within each block, timestep embeddings and conditioning features modulate the denoising. Self-attention captures dependencies among the noisy grasp tokens, while cross-attention injects geometric priors from the object and template hand.

Both $f_t$ and $\mathbf{c}$ are anchor-wise token sequences with a shared hidden dimension. Anchor positions $p_i$ are encoded via Rotary Positional Embeddings (RoPE)~\cite{su2024roformer} to preserve spatial layout; the remaining spectral features are projected by separate linear layers before entering the transformer blocks.

\paragraph{Neighborhood-Enhanced Attention}
Standard attention ignores geometric adjacency between neighboring patches. To exploit local structure, we add a pairwise attention bias $s_{ij}$ built from $k$-nearest neighbor graphs of template-hand and object patches. The induced anchor adjacencies form a sparse structural prior over token pairs, which is added to the attention logits:
\begin{equation*}
\alpha_{ij} = \mathrm{softmax}\left(q_i k_j^\top / \sqrt{d_h} + s_{ij}\right),
\end{equation*}

This bias encourages focus on structurally adjacent regions while preserving the long-range interactions needed for coherent hand-object reasoning.

\paragraph{Physics-Guided Training and Sampling}
Although diffusion captures geometric correlations between hand and object, geometry alone does not guarantee a physically plausible grasp. Following~\cite{zhong2025dexgrasp}, we incorporate physical guidance in both training and sampling.

We denote the combined physical penalty as $\mathcal{L}_{\mathrm{phys}}$, which is computed on point clouds reconstructed from generated InterMASH tokens. It serves as an auxiliary constraint during training to encourage physically plausible hand-object interactions, promoting stable contact while reducing penetration and self-collision. Full formulation and weighting are provided in the suppl. material.

For physics-guided sampling, the key step is to evaluate physical plausibility on a clean geometry estimate rather than directly on the noisy diffusion state. At denoising step $t$, given the current noisy token $f_t$ and the predicted noise $\epsilon_{\phi}(f_t,\mathbf{c},t)$, we estimate the corresponding denoised token as
\begin{equation*}
\hat{f}_0(f_t,t) = \frac{1}{\sqrt{\bar{\alpha}_t}}\left(f_t - \sqrt{1-\bar{\alpha}_t}\,\epsilon_{\phi}(f_t,\mathbf{c},t)\right).
\end{equation*}
We reconstruct hand and object point clouds from $\hat{f}_0$ and evaluate $\mathcal{L}_{\mathrm{phys}}(\hat{f}_0)$ on this denoised geometry. This estimate becomes more reliable at low-noise timesteps, where the generated geometry is close enough to the data manifold for physical penalties to provide meaningful gradients.

We then use the physical energy to guide the reverse diffusion trajectory. Let $\mu_{\phi}(f_t,\mathbf{c},t)$ and $\Sigma_t$ be the DDPM posterior mean and variance at step $t$. Instead of sampling from the unguided posterior mean, we shift it along the negative gradient of the physical penalty:
\begin{equation*}
    \widetilde{\mu}_{\phi} = \mu_{\phi}(f_t,\mathbf{c},t)
    - s\,\Sigma_t\nabla_{f_t}\mathcal{L}_{\mathrm{phys}}(\hat{f}_0(f_t,t)),
\end{equation*}
where $s$ is the guidance scale. This update keeps the sampler within the DDPM reverse process while steering each step toward lower penetration, fewer self-collisions, and more stable contact.

\subsection{Final Hand Pose from Inverse Kinematics}
\label{sec:ik}
The generative model outputs grasp-specific hand geometry in the InterMASH space, which we convert into an articulated pose $(\theta, R, \mathbf{p})$ through IK. We first reconstruct the generated hand as patch-wise point sets $\{\hat{\mathcal{S}}_i\}_{i=1}^{P}$ and initialize the global alignment $(R, \mathbf{p})$ from corresponding template and generated patch centers using closed-form Umeyama alignment~\cite{umeyama2002least}. The articulation parameters are then recovered by fitting the posed template-hand patches to the generated patches under the established anchor-patch correspondence. This patch-wise IK preserves anchor semantics and reduces matching ambiguity compared with global surface fitting. On generated ShadowHand samples from the filtered CMapDataset, patch-wise IK achieves an average total $\ell_1$ Chamfer error of 0.0125 m and a 99.5\% success rate under $\mathcal{L}_{\mathrm{CD}}^{\mathrm{total}}<0.1$ m, with an average decoding time of 0.39 seconds per sample. Detailed IK objectives, regularization terms, and optimization settings are provided in the supplementary material.

\section{Experiments}

We first describe the experimental setup (\autoref{sec:exp_setup}), and then evaluate our method from four perspectives: single-hand grasp generation to assess grasp quality and physical plausibility (\autoref{sec:exp_single}); cross-embodiment robotic grasp synthesis to test whether InterMASH supports joint modeling across morphologies (\autoref{sec:exp_cross}); human prior transfer to examine whether human grasp data can improve robotic grasp synthesis (\autoref{sec:exp_human}); and ablation studies to validate key design choices in our representation and generative model (\autoref{sec:exp_ablation}).

\subsection{Experimental Setup}
\label{sec:exp_setup}

\paragraph{Implementation Details}
Our model is implemented in PyTorch~\cite{paszke2019pytorch} and optimized using Muon~\cite{muon}. We follow official train-test splits and conduct all experiments on an Ubuntu server with 8 NVIDIA RTX 4090 GPUs.

\paragraph{Evaluation Metrics} We evaluate grasp quality, physical plausibility, and diversity. For single-hand grasp generation, we report \emph{Suc.6}, \emph{Suc.1}, \emph{Pen.}, and \emph{Div.} following the standard DexGraspNet~\cite{wang2022dexgraspnet} evaluation protocol. 
\emph{Suc.6} and \emph{Suc.1} denote success under all six disturbance directions and at least one disturbance direction, respectively.
\emph{Pen.} denotes the maximum penetration depth; and \emph{Div.} measures diversity as the mean standard deviation of local pose parameters. For cross-embodiment robotic grasp synthesis, we follow the evaluation protocol of $\mathcal{D(R,O)}$~\cite{wei2024d} and report \emph{Success Rate} and \emph{Diversity}. \emph{Success Rate} measures the percentage of generated grasps that remain stable after simulator-based execution and disturbance testing from six directions, while \emph{Diversity} measures the standard deviation of joint values among successful grasps.

\begin{table}[t]
    \caption{Performance comparison across different methods on DexGraspNet dataset. Bold numbers indicate the best scores, while underlined numbers indicate the second-best scores.}
    \centering
        \small
        \begin{tabular}{l|cccc}
            \toprule
            Method & \textbf{Suc.6 $\uparrow$} & \textbf{Suc.1 $\uparrow$} & \textbf{Pen. $\downarrow$} & \textbf{Div. $\uparrow$} \\ 
            \midrule
            UniDexGrasp (\citeyear{xu2023unidexgrasp})  & 33.9 & 70.1 & 31.9 & 0.14 \\
            GraspTTA (\citeyear{jiang2021graspTTA}) & 18.6 & 67.8 & 24.5 & 0.13 \\
            SceneDiffuser (\citeyear{huang2023diffusion})  & 26.6 & 66.9 & 31.0 & 0.15 \\
            UGG (\citeyear{lu2024ugg}) & 46.9 & 79.0 & 25.2 & 0.14 \\ 
            DGA (\citeyear{zhong2025dexgrasp}) & \textbf{53.6} & 90.4 & 21.5 & \textbf{0.22} \\
            $\mathcal{D(R,O)}$ (\citeyear{wei2024d}) & 46.9 & 89.7 & \underline{17.5} & \underline{0.20}  \\
            \midrule
            InterMASH (Ours) & \underline{53.5} & \textbf{91.9} & \textbf{16.2} & 0.14 \\
            \bottomrule
        \end{tabular}
    \label{tab:shadow_result}
\end{table}

\subsection{Single-Hand Grasp Generation}
\label{sec:exp_single}
To validate our framework's grasp generation for a single hand type, we evaluate on DexGraspNet~\cite{wang2022dexgraspnet}, a large-scale dataset for the five-fingered dexterous ShadowHand. 

\begin{figure*}
    \centering
    \includegraphics[width=0.9\linewidth]{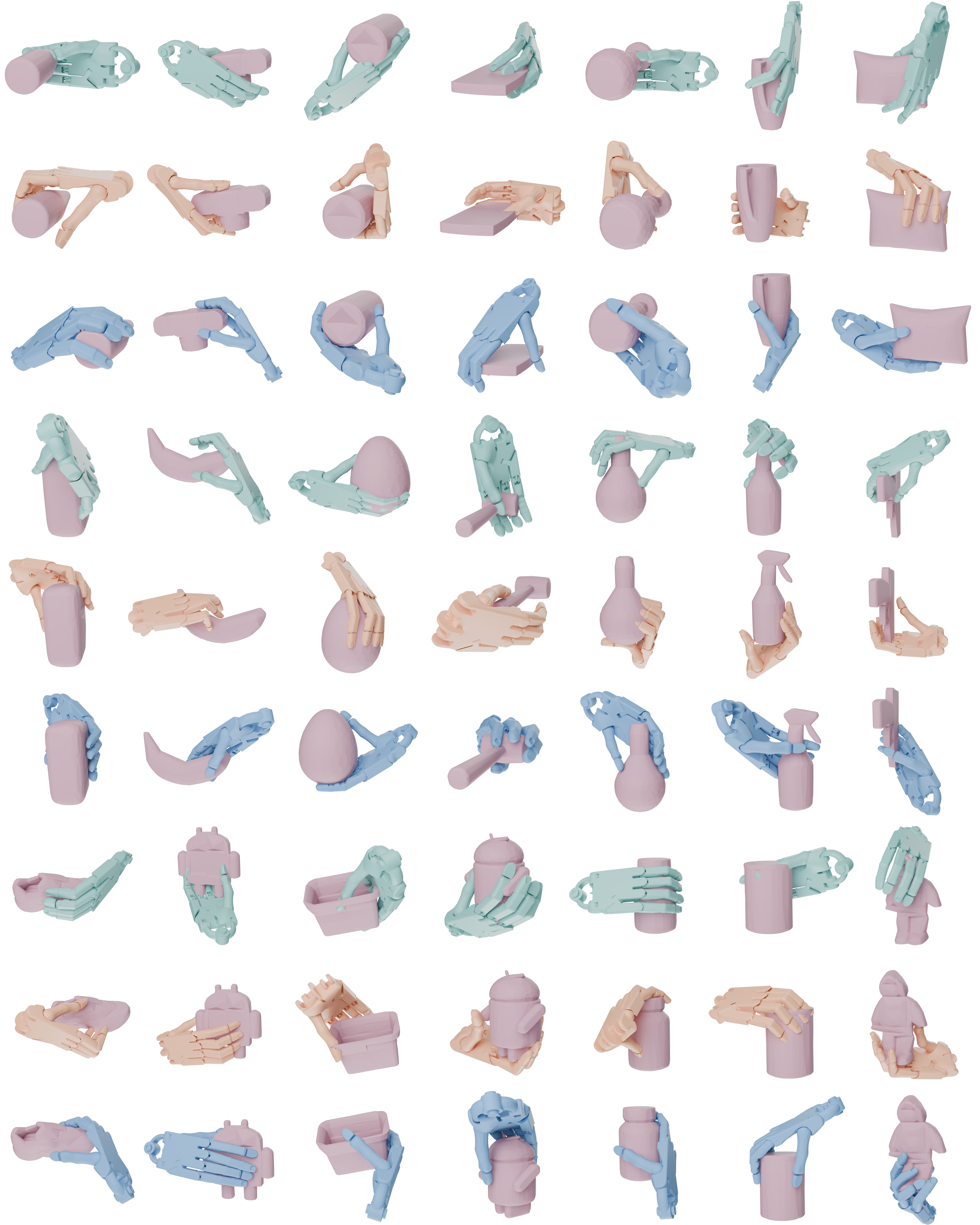}
    \caption{Qualitative comparison with baseline DexGrasp Anything~\cite{zhong2025dexgrasp} and $\mathcal{D(R,O)}$~\cite{wei2024d}. DexGrasp Anything's results are shown in green, $\mathcal{D(R,O)}$'s results are shown in orange, and our results are shown in blue.}
    \label{fig:gallery_shadow}
\end{figure*}
\begin{figure*}
    \centering
    \includegraphics[width=0.95\linewidth]{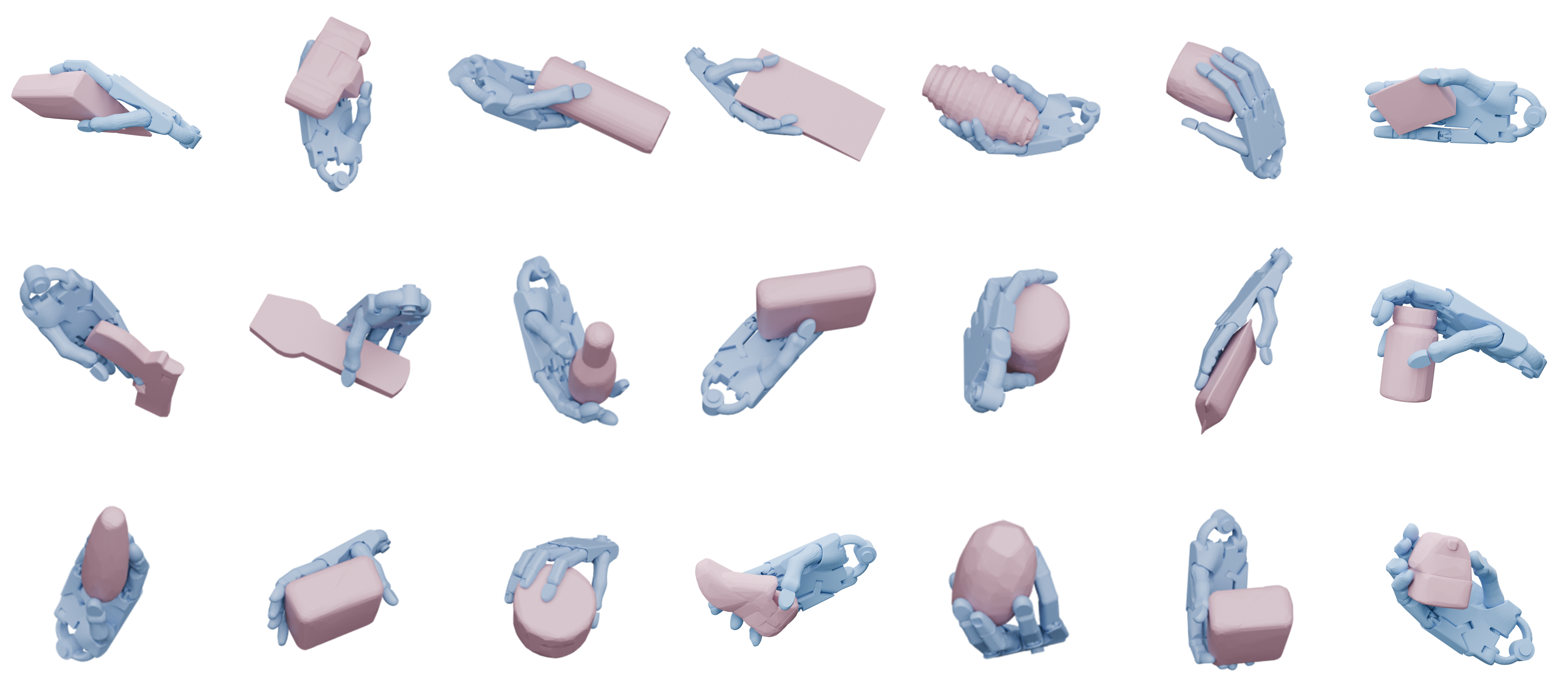}
    \caption{Visualization of our method's results on DexGraspNet~\cite{wang2022dexgraspnet} dataset. More results are shown in supplementary material.}
    \label{fig:shadow_results_part_1}
\end{figure*}

\autoref{tab:shadow_result} reports quantitative comparisons on DexGraspNet~\cite{wang2022dexgraspnet}; qualitative examples are shown in \autoref{fig:gallery_shadow} and \autoref{fig:shadow_results_part_1}. Our method achieves the best performance on \emph{Suc.1} and \emph{Pen.}, indicating a higher rate of grasps that remain stable under at least one disturbance direction and lower penetration. This aligns with our design: InterMASH jointly models local hand geometry and contact, while physics-guided training discourages implausible interactions.

On \emph{Suc.6} and \emph{Div.}, our method remains competitive but does not outperform the strongest baselines. This suggests a quality-diversity trade-off: the proposed representation and training objective bias the model toward more physically reliable grasps, at the cost of slightly lower diversity and performance under the stricter all-six-direction criterion.


\subsection{Cross-Embodiment Robotic Grasp Synthesis}
\label{sec:exp_cross}
To evaluate whether a shared InterMASH token space can jointly model multiple morphologies, we train a single model on the ShadowHand and Barrett subsets of the filtered CMapDataset~\cite{li2022gendexgrasp,wei2024d}. Following prior work, grasp success is evaluated in Isaac Gym~\cite{liang2018gpu} under the $\mathcal{D(R,O)}$ protocol; detailed evaluation procedures are provided in the Supplementary Material. We also measure diversity as the standard deviation of the joint angles in all successful grasps.

\begin{table}[t]
    \centering
    \caption{Cross-embodiment results and the effect of mixed-hand training on the filtered CMapDataset. Bold numbers indicate the best scores, while underlined numbers indicate the second-best scores.}
\resizebox{0.9\linewidth}{!}{
    \begin{tabular}{lcccc}
        \toprule
        \multirow{2}{*}{Method}
        & \multicolumn{2}{c}{Success Rate (\%) $\uparrow$}
        & \multicolumn{2}{c}{Diversity $\uparrow$} \\
        \cmidrule(lr){2-3} \cmidrule(lr){4-5}
        & Barrett & Shadow & Barrett & Shadow \\
        \midrule
        DFC (\citeyear{liu2021synthesizing})
            & 86.30 & 58.80
            & \textbf{0.532} & \underline{0.435} \\
        GenDexGrasp (\citeyear{li2022gendexgrasp})
            & 67.00 & 54.20
            & 0.488 & 0.318 \\
        $\mathcal{D(R,O)}$ (\citeyear{wei2024d})
            & \underline{87.30} & \textbf{83.00}
            & \underline{0.513} & \textbf{0.441} \\
        \midrule
        InterMASH (Shadow only)
            & -- & 57.62
            & -- & 0.416 \\
        InterMASH (Shadow+Barrett)
            & \textbf{90.30} & \underline{64.15}
            & 0.480 & 0.396 \\
        \bottomrule
    \end{tabular}
}
    \label{tab:mixed_results}
\end{table}


The quantitative results in \autoref{tab:mixed_results} show that, under a single model jointly trained on both Barrett and ShadowHand data, InterMASH achieves the highest success rate on Barrett while remaining competitive on ShadowHand. The improvement is more pronounced on Barrett, where the dataset is larger and the hand has fewer degrees of freedom (8 DoF). We also compare against a model trained only on ShadowHand data, and find that adding Barrett training data improves ShadowHand success from 57.62\% to 64.15\%, with only a small decrease in diversity from 0.416 to 0.396. This result suggests that mixed-hand training is beneficial.

\begin{figure}[t]
    \centering
    \includegraphics[width=0.9\linewidth]{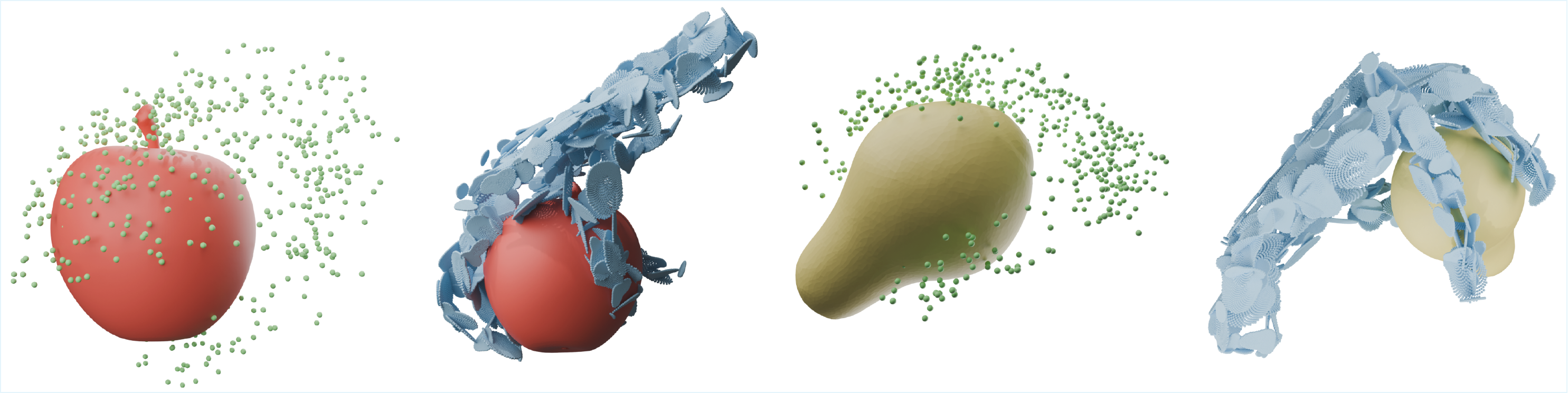}
    \caption{Qualitative comparison between grasp geometries generated by InterMASH and $\mathcal{D(R,O)}$.}
    \label{fig:dro_compare}
\end{figure}

\begin{figure*}
    \centering
    \includegraphics[width=0.9\linewidth]{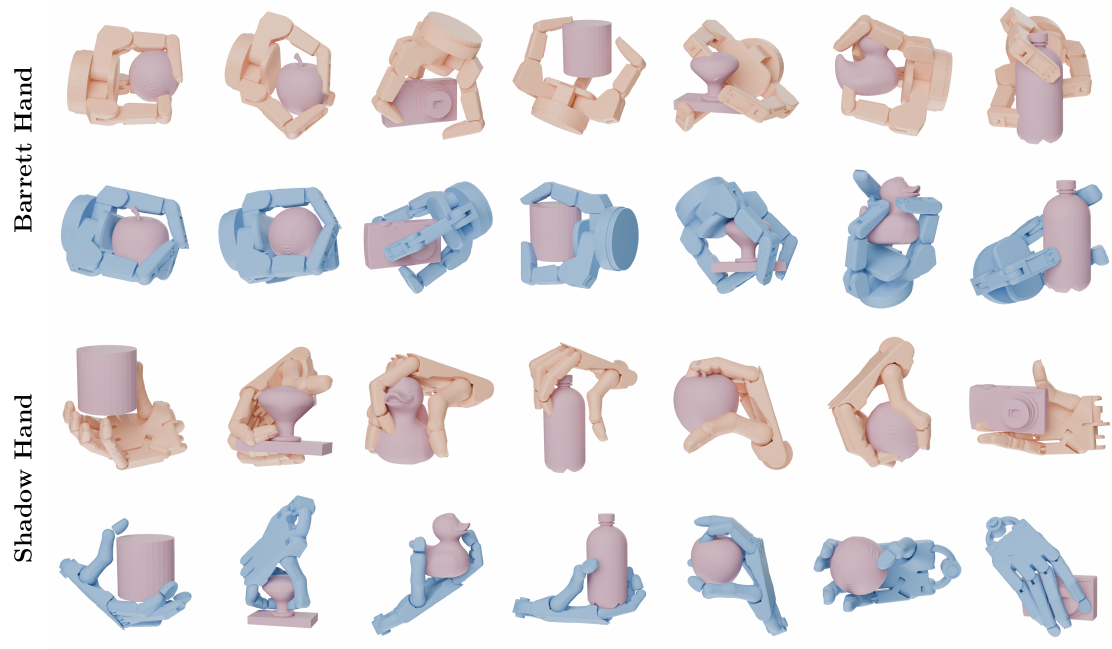}
    \caption{Qualitative comparison with $\mathcal{D(R,O)}$ on the filtered CMapDataset~\cite{wei2024d}. Baseline results are shown in yellow, and our results are shown in blue. The top two rows show Barrett Hand results, and the bottom two rows show Shadow Hand results. The visualization shows that our method tends to produce more plausible contact in these examples, whereas the baseline may exhibit insufficient contact or hand-object penetration.}
    \label{fig:gallery_barrett}
\end{figure*}

We also provide a qualitative comparison in \autoref{fig:dro_compare} to further illustrate the advantages of the InterMASH representation. The point clouds generated by $\mathcal{D(R,O)}$ are limited to a fixed resolution and do not clearly reveal hand structure, whereas InterMASH produces more recognizable hand-like geometry and supports arbitrary sampling through its continuous local patch representation. Additional cross-embodiment qualitative results are shown in \autoref{fig:gallery_barrett}.

\subsection{Integration of Human Priors into Robotic Grasping}
\label{sec:exp_human}

To study whether InterMASH can incorporate human grasp priors into robotic grasp synthesis, we compare a baseline model trained on DexGRAB~\cite{zhong2025dexgrasp} with a counterpart fine-tuned on a 1:1 mixture of DexGRAB and MANO-based GRAB~\cite{GRAB-2020} samples. Both models are evaluated on the same DexGRAB test split.

As shown in \autoref{tab:human_prior_transfer}, GRAB fine-tuning improves success and diversity, indicating beneficial human-to-robot prior transfer. Penetration increases from 14.9 to 18.6 mm, likely because GRAB training data have higher average penetration than DexGRAB (24.6 vs.\ 11.34 mm), showing that transfer quality also depends on source-data geometry.

\begin{table}[t]
    \centering
    \small
    \caption{Effect of integrating human priors. The table compares two training settings: using only ShadowHand DexGRAB samples, and using a mixture of ShadowHand DexGRAB and human-hand GRAB samples. Bold numbers indicate the best scores.}
    \small
    \begin{tabular}{l|cccc}
        \toprule
        Training Setting & \textbf{Suc.6 $\uparrow$} & \textbf{Suc.1 $\uparrow$} & \textbf{Pen. $\downarrow$} & \textbf{Div. $\uparrow$} \\ 
        \midrule
        DexGRAB only & 25.8 & 64.2 & \textbf{14.9} & 0.450 \\
        DexGRAB+GRAB fine-tune & \textbf{29.0} & \textbf{65.9} & 18.6 & \textbf{0.500} \\
        \bottomrule
    \end{tabular}
    \label{tab:human_prior_transfer}
\end{table}

\subsection{Ablation Study}
\label{sec:exp_ablation}

We conduct ablation studies to analyze the contribution of the key components introduced in \autoref{sec:method}. Specifically, under mixed-hand training setting, we evaluate effects of interaction representation choice, physics-guided training, physics-guided sampling, and Neighborhood Enhanced Attention by selectively enabling or disabling each component.

\paragraph{Ablation on Interaction Representations} To isolate the effect of representation, we compare InterMASH with BPS~\cite{prokudin2019efficient} and VAE-based patch embeddings under a matched DiT and IK pipeline. All variants use 128 anchors/patches, while physics-guided training, physics-guided sampling, and Neighborhood Enhanced Attention are disabled to ensure a fair comparison. As shown in \autoref{tab:ablation_repr}, InterMASH consistently achieves the highest success rate across all hand types, showing the benefit of combining fixed anchor-wise correspondence with compact local patch descriptors and object-side contact. More details are provided in suppl. material.


\begin{table}[t]
    \centering
    \caption{Representation ablation under matched generation settings on the filtered CMapDataset~\cite{li2022gendexgrasp,wei2024d}. Bold numbers indicate the best scores.}
    \resizebox{0.9\linewidth}{!}{
    \begin{tabular}{lccc}
        \toprule
        \multirow{2}{*}{Method}
        & \multicolumn{3}{c}{Success Rate (\%) $\uparrow$} \\
        \cmidrule(lr){2-4}
        & Barrett & Allegro & Shadow \\
        \midrule
        Basis Point Set & 4.50 & 3.90 & 5.40 \\
        Patch Emb. + Soft-Intro VAE & 33.20 & 11.10 & 14.30 \\
        Patch Emb. + In-house VAE & 24.80 & 11.10 & 10.40 \\
        InterMASH (Ours) & \textbf{52.70} & \textbf{15.90} & \textbf{20.00} \\
        \bottomrule
    \end{tabular}
}
    \label{tab:ablation_repr}
\end{table}

\paragraph{Ablation on Key Components in InterMASH}
\autoref{tab:ablation_key_component} reports the effect of the main components in our method. Without physics-guided training or Neighborhood Enhanced Attention, the model attains relatively high diversity but noticeably lower success rates. Adding physics-guided training substantially improves success on both Barrett and ShadowHand, confirming the importance of explicit physical supervision for grasp synthesis. Physics-guided sampling further improves ShadowHand success, and enabling Neighborhood Enhanced Attention gives the strongest overall performance. The moderate drop in diversity is consistent with the stronger geometric and physical constraints imposed by these components.


\begin{table}[t]
    \centering
    \caption{Ablation study on key components of our method. Bold numbers indicate the best scores.}
    \resizebox{0.9\linewidth}{!}{
    \begin{tabular}{ccccccc}
        \toprule
        \multirow{2}{*}{Phys. Train} &
        \multirow{2}{*}{Phys. Samp.} &
        \multirow{2}{*}{N.E.A.} &
        \multicolumn{2}{c}{Suc.\,$\uparrow$} &
        \multicolumn{2}{c}{Div.\,$\uparrow$} \\
        \cmidrule(lr){4-5} \cmidrule(lr){6-7}
        & & & Barrett & Shadow & Barrett & Shadow \\
        \midrule
        \xmark  & \xmark  & \xmark
            & 80.17 & 51.06 & \textbf{0.498} & \textbf{0.430} \\
        \cmark & \xmark  & \xmark
            & 86.10 & 59.50 & 0.496 & 0.426 \\
        \cmark & \cmark & \xmark
            & 85.20 & 61.20 & 0.496 & 0.427 \\
        \midrule
        \cmark & \cmark & \cmark
            & \textbf{90.30} & \textbf{64.15} & 0.480 & 0.396 \\
        \bottomrule
    \end{tabular}
    }
    \label{tab:ablation_key_component}
\end{table}

\section{Conclusion}
\label{sec:conclusion}

We presented InterMASH, a unified geometric representation for hand--object interaction based on anchor-indexed patch-level primitives. Built on this representation, we introduced a conditional diffusion model that jointly generates grasping hand geometry and object-side contact conditioned on object geometry and a hand template. Experiments show that InterMASH improves physical feasibility and contact consistency, supports cross-embodiment robotic hand training, and enables the integration of MANO-based human grasp data into robotic grasp synthesis. These results suggest that intermediate geometric interaction representations are an effective design choice for scalable grasp generation.

\paragraph{Limitation and Future Work}
A current limitation is that InterMASH uses a relatively general anchor-indexed representation with fewer hand-specific inductive biases than methods such as $\mathcal{D(R,O)}$. This design improves scalability and performs strongly when sufficient data are available, as shown by the DexGraspNet results, but it can be less data-efficient for high-DoF hands in smaller mixed-hand datasets. For example, in the filtered CMapDataset setting, ShadowHand has fewer samples and a higher-dimensional action space than Barrett, where stronger built-in correspondence priors can still be advantageous. Future work could combine the scalability of InterMASH with additional data-efficient priors, pretraining objectives, or hand-aware conditioning for high-DoF embodiments. We discuss a separate handedness ambiguity observed in mixed left- and right-handed robotic training in the supplementary material.


\begin{acks}
This work is financially supported by Laoshan Laboratory  (No. LSKJ202300305), by the National Natural Science Foundation of China (NSFC) under Grants 62325211 and 62132021, and by the Research Fund of the Jiangsu Key Laboratory of AI for Industries under Grant E6420016G8. The numerical calculations in this paper have been done on the supercomputing system in the Supercomputing Center of University of Science and Technology of China.
\end{acks}

\clearpage
\onecolumn
\twocolumn
\bibliographystyle{ACM-Reference-Format}
\bibliography{main}

\end{document}